# Dermatologist Level Dermoscopy Skin Cancer Classification Using Different Deep Learning Convolutional Neural Networks Algorithms

Amirreza Rezvantalab[1]*, Habib Safigholi[2]*, Somayeh Karimijeshni[3]

1-Department of Computer Engineering, Shiraz branch, Islamic Azad University, Shiraz, Iran

2-Department of Electrical Engineering, Shiraz branch, Islamic Azad University, Shiraz, Iran

3-Legal Medicine Research Center, Iranian Legal Medicine Organization, Tehran, Iran

**Abstract**

In this paper, the effectiveness and capability of convolutional neural networks have been studied in the classification of 8 skin diseases. Different pre-trained state-of-the-art architectures (DenseNet 201, ResNet 152, Inception v3, InceptionResNet v2) were used and applied on 10135 dermoscopy skin images in total (*HAM10000*: 10015, *PH$^2$*: 120). The utilized dataset includes 8 diagnostic categories - melanoma, melanocytic nevi, basal cell carcinoma, benign keratosis, actinic keratosis and intraepithelial carcinoma, dermatofibroma, vascular lesions, and atypical nevi. The aim is to compare the ability of deep learning with the performance of highly trained dermatologists. Overall, the mean results show that all deep learning models outperformed dermatologists (at least 11%). The best ROC AUC values for melanoma and basal cell carcinoma are 94.40% (ResNet 152) and 99.30% (DenseNet 201) versus 82.26% and 88.82% of dermatologists, respectively. Also, DenseNet 201 had the highest macro and micro averaged AUC values for overall classification (98.16%, 98.79%, respectively).

## 1. Introduction

Skin cancer is the most widespread cancer diagnosed in the US [1]. Melanoma is the most dangerous type of skin cancer which has been one of the most important challenges of the public health in recent years [2].

According to the latest statistics, 91270 new cases of melanoma is predicted to be diagnosed in the United States in 2018 [1]. The rates of melanoma occurrence and the mortality result of this disease, is expected to rise over the next decades [3]. Recent report shows that from 2008 to 2018, there has been a 53% increase in new melanoma case diagnosis annually [1, 4]. If this type of cancer can be diagnosed in its early phases, with choosing the appropriate treatment [5, 6], survival rates are very promising [7]. Otherwise, the predicted 5-year survival rate of a patient will be reduced from 99% to 14% [8-10]. Also in non-melanoma type of cancer, there has been a drastic increase in the diagnosis of new cases up to 77% between 1994 and 2014. The basal cell carcinoma is the most common type of non-melanoma skin cancer which results in the death of 3000 people each year [11].

So, this will cause a high demand for primary inspection and detection of different skin cancer types in order to prevent it from getting worse and give a chance for better prognosis [3]. The usually way for a clinician to detect melanoma is by inspecting the visual details which has a low precision [12, 13]. On the other hand, dermoscopy [14] is a non-invasive technique that can capture a high resolution image of the skin which enables dermatologists to detect features which are invisible to the naked eye.

By several meta-analyses, it has been shown that with the use of dermoscopy, there will be an improvement and a higher accuracy in the diagnosis of melanoma compared to the naked eye inspection [15-17]. However, this technique in the hands of an unskilled clinician would results in a poor performance [18-21], not to mention that even for experts, this task is time-consuming and based on their judgement and therefore highly subjective and they may produce even different diagnostic results [18, 22, 23]. Because of the resemblance between malignant skin tumors and benign skin lesions in visual features, it is very hard for dermatologists to differentiate between them. The average reported sensitivity for melanoma detection is often less than 80% even with highly trained dermatologists and clinicians [24, 25]. Moreover, highly trained dermatologists are not available all over the world. Therefore, automatic skin lesion classification at the same level and accuracy as dermatologists or even higher is very crucial in public health.

In order to solve aforementioned problems, there has been extensive publications on classifying malignant and non-malignant skin cancer by developing computer image analysis algorithms. These algorithms use a variety of approaches towards the segmentation, detection and classification of melanoma by integrating areas like image processing, computer vision and machine learning [20, 26-29]. The main problems of these works are insufficient data and the lack of diversity in skin cancer classes. Also in [30-33], different approaches were used to classify melanoma, but these approaches rely heavily on 'man-made' segmentation criteria which is a limitation.

Almansour et al. [34] developed an algorithm for the classification of melanoma by using k-means clustering, and Support Vector Machine (SVM). Abbas et al. [35], and Capdehourat et al. [36] separately used AdaBoost MC to classify skin lesions. Giotis et al. [37], and Ruiz et al. [10] used different set of features (lesion texture, visual, color, etc.) and neural networks for the developing a decision support system. Also, Isasi et al. [38] developed an algorithm for diagnosis of melanoma. Past research on machine learning methods and their applications in pattern recognition systems was restricted to the transformation of a raw input into a representation of important hand-made feature vectors which then, could be fed to a classifier for the classification and detection of patterns. However in the past recent years with the exponential raise in the computation power and a large amount of data, deep learning techniques has become popular among researchers. These powerful methods use representation learning which enables the model to learn the representation of the raw data hierarchically in multiple layers with several abstractions [39]. The advantage of deep learning is that it can learn features directly from data without the help of any human expert for feature engineering and can exceed human performance.

Recently, deep learning algorithms has gained great success in different computer vision problems. In 2012, Krizhevsky et al. [40] developed a novel technique (AlexNet) by using convolutional neural networks for the classification of a big data (1.2 million images) containing 1000 object categories in the ImageNet Large Scale Visual Recognition Challenge 2010 (ILSVRC2010) and achieved the best result and therefore an enormous attention among academics in computer vision area. In [9], Esteva et al. made a breakthrough on skin cancer classification by a pre-trained GoogleNet Inception v3 CNN model to classify 129,450 clinical skin cancer images including 3,374 dermatoscopic images. Yu et al. [41] developed a convolutional neural network with over 50 layers on ISBI 2016 challenge dataset for the classification of malignant melanoma. In 2018, Haenssle et al. [42] utilized a deep convolutional neural network to classify a binary diagnostic category of dermoscopy images of melanocytic images. Dorj et al. [8] developed ECOC SVM with deep convolutional neural network approach for the classification of 4 diagnostic categories of clinical skin cancer images. Han et al. [43] used a deep convolutional neural network to classify the clinical images of 12 skin diseases. To our best knowledge in the previous studies, we did not find any research on the use of dermoscopy skin cancer dataset with enough diversity for multi-class classification of dermatoscopic skin cancer images.

This paper aims at testing a deep learning approach for a multi-class classification with 8 major diagnostic categories by applying state-of-the-art pre-trained deep convolutional neural networks on 2 public dermoscopy skin cancer databases which can yield a higher diagnostic accuracy compared to dermatologists. This approach will help clinicians for better diagnosis across different skin cancer categories. Since there are too many individual smartphones in the world, one can therefore developed some potentially low-cost universal access apps based on training deep learning approach to visually screen the skin cancer at the early stage, to help define the best treatment. The rest of paper is arranged as follows: Section 2 describes methods and materials. Section 3 represents experimental results. In Section 4 and 5, discussion and conclusion are presented.

## 2. Methods and materials

### 2.1. Materials

In this research, the aim is to differentiate between not only melanoma and nevus, but also non-melanocytic pigmented lesions which are common in practice. Therefore, a combination of two public datasets is utilized in this paper to classify different skin cancer images by pre-trained convolutional neural networks. The first one is the *HAM10000* dataset [44] which is served as a benchmark database for academic machine learning purposes. We used the training set part of this dataset which consists of 10015 dermatoscopic images with the size of $450 \times 600$ and includes 7 diagnostic categories as follows: melanoma (Mel, 1113 samples), melanocytic nevi (NV, 6705 samples), basal cell carcinoma (BCC, 514 samples), actinic keratosis and intraepithelial carcinoma

(AKIEC, 327 samples), benign keratosis (BKL, 1099 samples), dermatofibroma (DF, 115 samples), vascular lesions (VASC, 142 samples).

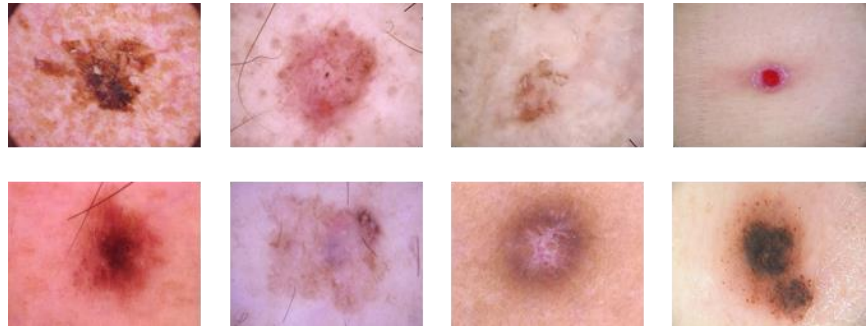

**Figure. 1** Examples of the training dataset images. First row from left to right: melanoma, basal cell carcinoma, beningn keratosis, vascular lesions. Second row from left to right: melanocytic nevi, AKIEC, dermatofibroma, Atypical nevi.

The second dataset is the *PH$^2$* [45]. This database contains of 200 dermatoscopic images of melanocytic skin lesions. They are 8-bit RGB color images with a resolution of 768 × 560 pixels which we used images of atypical nevi (Atyp NV, 80 samples) and melanoma (Mel, 40 samples) categories of this database to expand our diagnostic categories and therefore, added up the entire diagnostic classes of dataset to 8 for this paper. Examples of the training dataset images are shown in Figure 1.

## 2.2. Methods

In over past few years, more advanced convolutional neural networks have been developed by researchers in order to solve computer vision problems more accurately. In this paper to classify skin cancer images, we utilized 4 deep convolutional neural networks pre-trained on ImageNet [46] using TensorFlow [47] which is a deep learning framework developed by Google. We split 70% of the dataset as training set, 15% as validation set and 15% as testing set to evaluate 4 models. Figure 2 shows us the architectures of 4 models. Fine-tuning every model is described in details as follows:

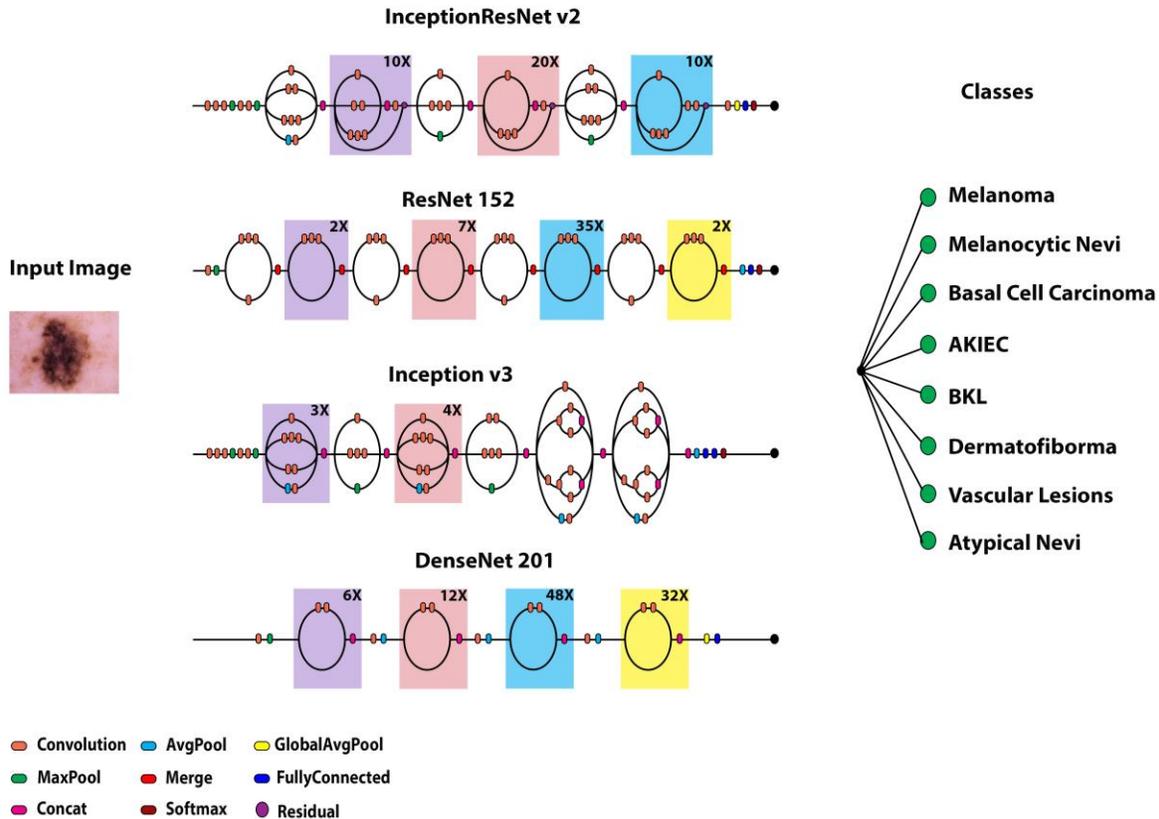

**Figure 2.** Architectures of 4 models

### 2.2.1. Google's Inception v3

Google's Inception v3 architecture [48] was re-trained on our dataset by fine-tuning across all layers and replacing top layers with one averagepooling, two fully connected and finally the softmax layer allowing to classify 8 diagnostic categories. The size of input images was all resized to (299, 299) to be compatible with this model. Learning rate was set to 0.0007 and stochastic gradient descent algorithm (SGD) with a decay and momentum of 0.9 was used for the optimizer.

### 2.2.2. InceptionResNet v2

InceptionResNet v2 architecture [49] was re-trained on our dataset by fine-tuning across all layers and replacing top layers with one globalaveragepooling, one fully connected and finally the softmax layer allowing to classify 8 diagnostic categories. The size of input images was all resized to (224, 224) to be compatible with this model. Learning rate was set to 0.0006 and stochastic gradient descent algorithm (SGD) with a decay and momentum of 0.9 was used for the optimizer.

### 2.2.3. ResNet 152

ResNet 152 architecture [50] was re-trained on our dataset by fine-tuning across all layers and replacing top layers with one averagepooling, one fully connected and finally the softmax layer allowing to classify 8 diagnostic categories. The size of input images was all resized to (224, 224) to be compatible with this model. Learning rate was set to 0.0006 and stochastic gradient descent algorithm (SGD) with a decay and momentum of 0.9 was used for the optimizer.

2.2.4. DenseNet 201

DenseNet 201 architecture [51] was re-trained on our dataset by fine-tuning across all layers and replacing top layers with one globalaveragepooling and the softmax layer allowing to classify 8 diagnostic categories. The size of input images was all resized to (224, 224) to be compatible with this model. Learning rate was set to 0.0006 and stochastic gradient descent algorithm (SGD) with a decay and momentum of 0.9 was used for the optimizer.

## 3. Results

In this section, we present our findings and show the diagnostic accuracy of trained models in comparison to an approved board of dermatologists. In this paper, a total of eight different types of major skin cancer categories is used. The evaluation and results of trained models is calculated by common classification metrics which are defined as follows:

$$Precision\ (positive\ predictive\ value) = \frac{TP}{(TP+FP)} \qquad (1)$$

$$Recall\ (true\ positive\ rate) = \frac{TP}{(TP+FN)} \qquad (2)$$

$$F_1\text{-score} = \frac{2 \times TP}{2 \times TP + FP + FN} \qquad (3)$$

Where TP is the number of positive cases which are labeled correctly, TN is the number of negative cases which are labeled correctly, FP is the number of positive cases which are labeled falsely, and FN is the number of negative cases which are labeled falsely. Also since the distribution of the number of samples among database are highly unbalanced, $F_1$-score result which is the harmonic mean of precision and recall, is reported and in order to have a graphical view of the trade-off between sensitivity and specificity metrics, ROC curves and their associated AUC values are used.

The performance of fined-tuned models is shown in Figure 3. This figure represents the area under curve (AUC) of receiver operating characteristic (ROC) of each category for different models. Moreover, because the dataset is unbalanced, we calculated macro and micro averaged ROC AUC scores. For DenseNet 201, the macro average ROC AUC is 98.16% and micro average ROC AUC is 98.79% which are the highest AUC scores among all models.

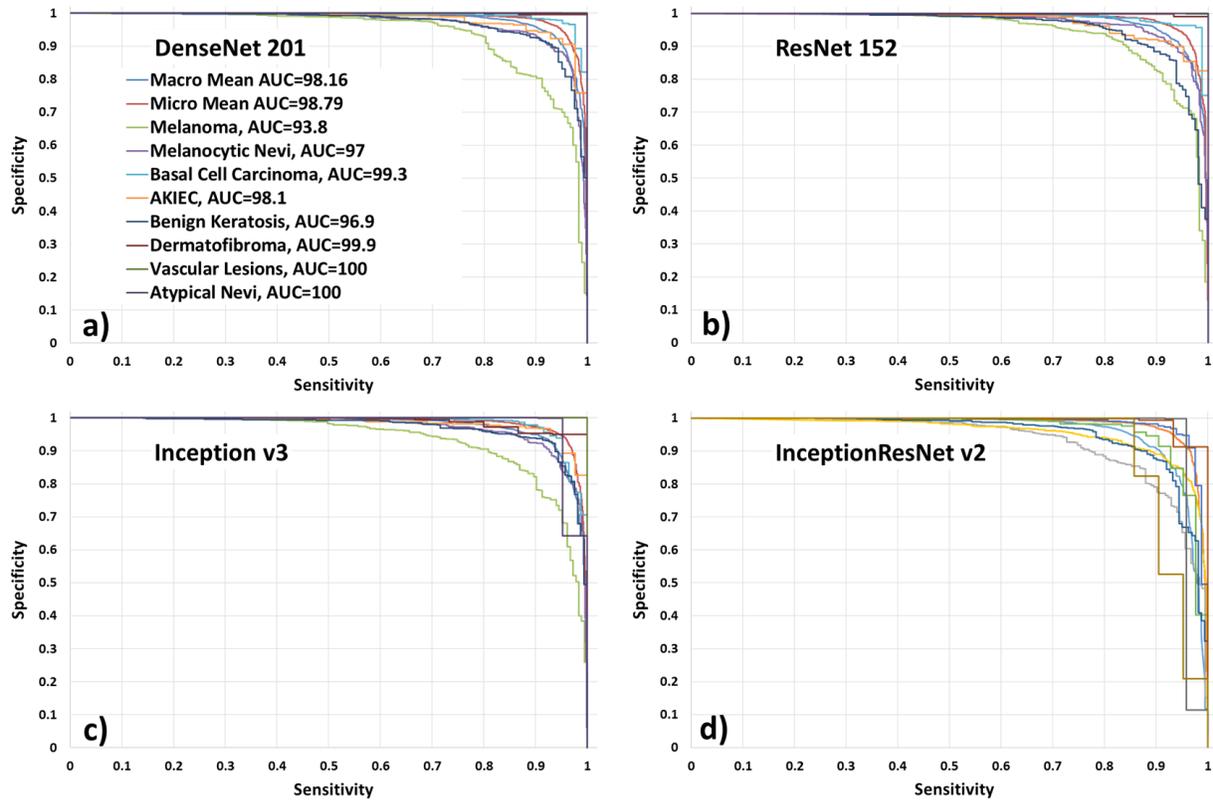

**Figure 3.** ROC curves and AUC values for each and overall diagnostic category of (a) DenseNet 201. (b) ResNet 152. (c) Inception v3. (d) InceptionResNet v2

Table 1 presents the details of ROC AUC for each algorithm. It can be seen that except for Benign keratosis and AKIEC, DenseNet 201 and ResNet 152 algorithms can achieve the highest AUC scores among the other models for each individual diagnostic category, especially for cancer categories like melanoma and basal cell carcinoma.

Table 2 summarizes the comparison of 4 models by reporting micro and macro average of precision, F1-score and ROC AUC metrics. The precision shows the capability of a classifier not to label as positive a sample that is negative. $F_1$-score reaches its best value near 1 and conversely. This shows us that DenseNet 201 has achieved better results overall, against other models in the classification of dermoscopy skin cancer images.

|  | ROC AUC (%) | | | | | | | | | |
|---|---|---|---|---|---|---|---|---|---|---|
| **Algorithm** | Mel | NV | BCC | AKIEC | BK | DF | VASC | Atyp NV | Macro | Micro |
| **DenseNet 210** | 93.8 | 97 | **99.3** | 98.1 | 96.9 | **99.9** | 100 | 100 | **98.16** | **98.79** |
| **ResNet 152** | **94.4** | **97.3** | 99.1 | 97.6 | 95.8 | 99.8 | 100 | 100 | 98.04 | 98.61 |
| **Inception v3** | 93.4 | 97 | 98.6 | **98.4** | **97.1** | 99 | 100 | 98.3 | 97.80 | 98.60 |
| **InceptionResNet v2** | 93.2 | 95.7 | 98.6 | 96.8 | 95.2 | 99.3 | 96.3 | 93.1 | 96.10 | 98.20 |

**Table 1.** ROC AUC values (%) for every algorithm on each and overall diagnostic category

| **Classifier** | **Precision (%)** **Micro - Macro** | **$F_1$ - score (%)** **Micro - Macro** | **ROC AUC (%)** **Micro - Macro** |
|---|---|---|---|
| **DenseNet 201** | **89.01 - 85.24** | **89.01 - 85.13** | **98.79 - 98.16** |
| **ResNet 152** | 88.22 - 81.29 | 88.22 - 82.09 | 98.61 - 98.04 |
| **Inception v3** | 86.84 - 80.22 | 86.84 - 79.60 | 98.60 - 97.80 |
| **InceptionResNet v2** | 86.90 - 83.37 | 86.90 - 81.18 | 98.20 - 96.10 |

**Table 2.** Micro and Macro averaged Precision, $F_1$ - score, and ROC AUC for each algorithm

Moreover, confusion matrices of each algorithm is presented in Figure 4. On the main diagonal of each matrix, we have the recall value of each related diagnostic category. It can be concluded that all models confuse melanoma and melanocytic nevi together. In the same manner, All 4 models did a poor job on the classification of AKIEC and benign keratosis categories. However, DenseNet 201 and ResNet 152 have achieved better results overall.

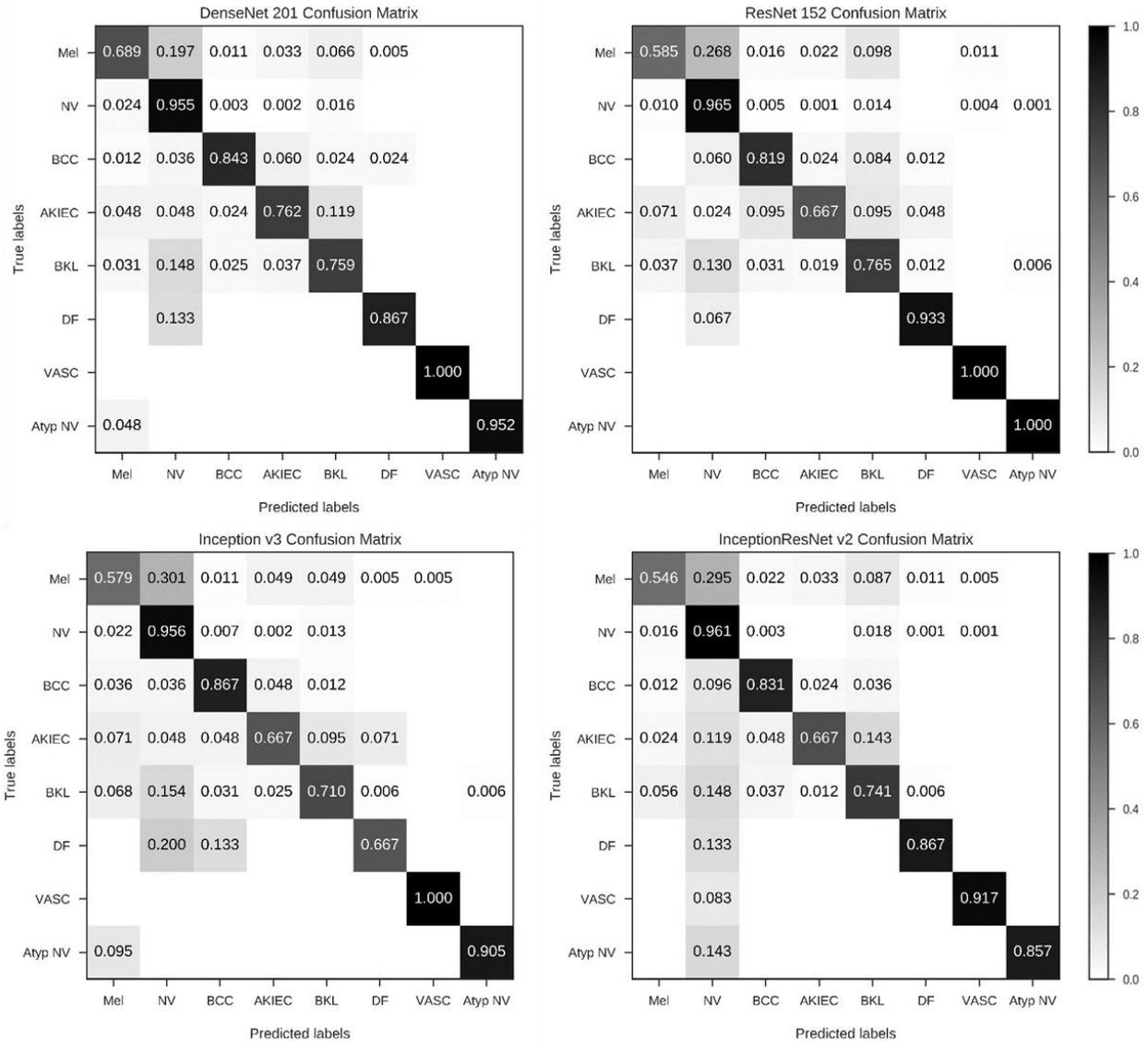

**Figure 4.** Confusion matrix of each model

In the context of the comparison of CNN's accuracy with the performance of dermatologists on cancer cases, we got interesting results that indicates the superior diagnostic accuracy of deep convolutional neural networks over expert clinicians. Figure 4 depicts the comparative results of ROC curves for each model versus dermatologists on both melanoma and basal cell carcinoma conditions.

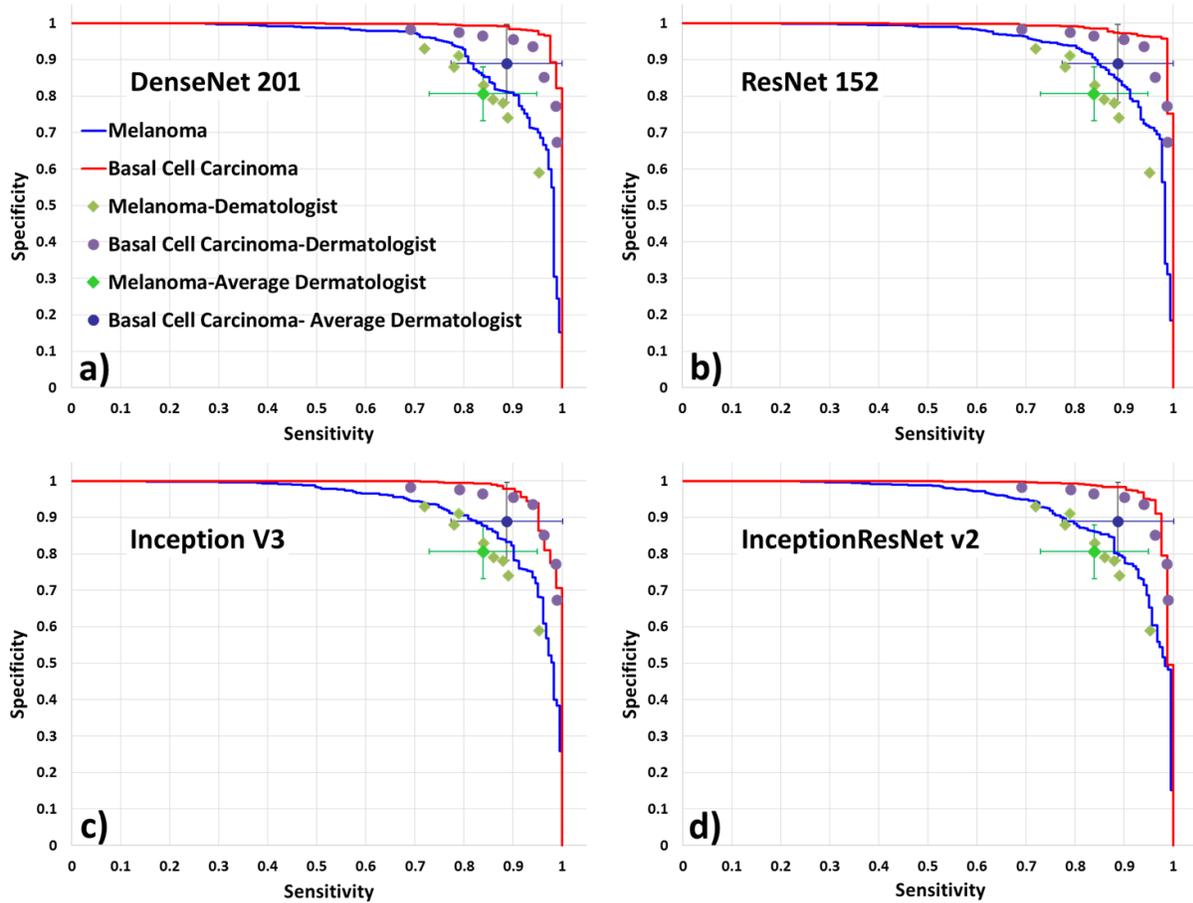

**Figure 3.** ROC curves for melanoma and basal cell carcinoma cases (blue, red) and the mean AUC values (blue circle) of (a) DenseNet 201. (b) ResNet 152. (c) Inception v3. (d) InceptionResNet v2 in comparison to the mean AUC of all dermatologists (mean: green circle)

In each comparative curve, dermatologists' diagnostic accuracy on cancer cases are the criterion in which the CNN's accuracy is compared to. Dermatologists' ROC AUC is based on the averaged sensitivity and specify score of dermatologists. The performance of each individual expert is also depicted. It can be seen that in both melanoma and basal cell carcinoma conditions, all of the algorithms have much greater ROC AUC than expert dermatologists. For Melanoma case, the highest ROC AUC is achieved by ResNet 152 (94.4%) in comparison to the ROC AUC of dermatologist (82.26%). Similarly, in Basal cell carcinoma case the highest ROC AUC is achieved by DenseNet 201 (99.3%) versus dermatologists' performance (88.82%). The sensitivity and specify details of each algorithm and dermatologists are summarized in Table 3.

| Classifier | ROC AUC (%) (Mel - BCC) |
|---|---|
| **Dermatologist** | 82.26 - 88.82 |
| **DenseNet 201** | 93.80 - **99.30** |
| **ResNet 152** | **94.40** - 99.10 |
| **Inception v3** | 93.40 - 98.60 |
| **InceptionResNet v2** | 93.20 - 98.60 |

**Table 3.** The comparative ROC AUC results (%) between dermatologists and each algorithm on cancer cases

## 4. Discussion

Skin cancer disease has been the most diagnosed cancer in the U.S each year [1, 3]. In recent years with the great success of deep convolutional neural networks in AI community, there has been made many efforts to analyze medical images [52-54]. To investigate this purpose further, we demonstrate the capability of CNNs to diagnose varied types of dermoscopy skin lesion images. This study, presents a deep learning approach for fully automated analysis of dermatoscopic images of skin diseases. Our aim was to implement state-of-the-art CNN architectures in order to test their ability in analyzing dermatoscopic images of skin and comparing their performance against expert dermatologists. Using a combination of *HAM10000* and *PH$^2$* datasets, four CNN models is used for the multi-class classification task (8 class). The ROC AUC values for each category is as follows: For basal cell carcinoma and dermatofibroma cases, DenseNet 201 achieved the highest AUC with the values of 99.3% and 99.9% respectively. For melanoma and melanocytic nevi classes, ResNet 152 scored the best with 94.4% and 97.3% respectively. For atypical nevi and vascular lesions categories, both DenseNet 201 and ResNet 152 achieved 100%. Also for AKIEC and benign keratosis classes, Google's Inception v3 scored the highest ROC AUC values, 98.4% and 97.1% respectively. Within the comparison of CNNs classification accuracy to the performance of dermatologists, experimental results shows the higher ROC AUC of CNNs over experts. In the classification of cancer categories, highly experienced dermatologists' ROC AUC values are 82.26% for melanoma and 88.82% for basal cell carcinoma. Whereas, all of the trained models, diagnosed melanoma and basal cell carcinoma with a superior ROC AUC compared to experts as follows: For melanoma, the highest ROC AUC is achieved by ResNet 152 (94.40%) while InceptionResNet v2 scored 93.20% as the worst among the other models. Similarly, for basal cell carcinoma DenseNet 201 has the best ROC AUC (99.30%) while the other models achieved 98.60% and 99.10%. As is shown in Figure 5, each model had a greater AUC compared to each individual expert. Also, DenseNet 201 has the best accuracy among the other models in terms of micro and macro averaged precision (89.01% - 85.24%), $F_1$ - Score (89.01% - 85.13%), and ROC AUC (98.79% - 98.16%).

Finally, we have compared our findings with some of the latest works on skin cancer classification. However as far as we know, past works on multi-class classification of skin diseases were reported mostly on only clinical images and this is the first research on a multi-class classification of only dermoscopy images. Moreover, the utilized dataset of this research is different from past research. But here we present some points about ROC AUC values of the latest works on using deep learning to classify skin cancer images. Yu et al. tried to apply deep CNNs like GoogleNet, VGG-116, and DRN-50 on ISBI 2016 challenge dataset in order to classify melanoma condition and for ROC AUC metric, best result was achieved with VGG-16 (82.6%). Esteva et al. used Google's Inception v3 on 129,450 images including 3374 dermoscopy images for binary classification (malignant melanomas versus benign nevi) and compared it with the performance of expert dermatologists. For the first time they showed that artificial intelligence has the ability to diagnose deadliest types of skin cancers with the same level of dermatologists. For melanoma they reported a ROC AUC of 91% for 111 dermoscopy images which was higher than the average of dermatologists. Han et al. applied ResNet 152 on 19,398 clinical images to classify 12 major skin disease categories. The AUC of melanoma and basal cell carcinoma classes with the Asan dataset, are 96 ± 0.01% and 96% respectively. While with the Edinburgh dataset, the AUC of mentioned classes are 90% and 88% respectively. Haenssle et al. trained the Google's Inception v4 architecture on ISBI 2016 challenge dataset for the classification of melanoma and achieved ROC AUC of 86% versus 79% and 82% of dermatologists.

## 5. Conclusion and Future works

In conclusion, this study investigated the ability of deep convolutional neural networks in the classification of 8 major skin diseases. Our results show that state-of-the-art deep learning architectures trained on dermoscopy images (10135 in total) outperforms dermatologists. We showed that with use of very deep convolutional neural networks and fine-tuning them on dermoscopy images, better diagnostic accuracy can be achieved compared to expert physicians and clinicians. Although, the utilized dataset is highly unbalanced and also no preprocessing step is applied in this paper, but the experimental results are very promising. These models can be easily implemented in dermoscopy systems or even on smartphones in order to assist dermatologists. More diverse datasets (varied categories, different ages) with much more dermoscopy images and balanced samples per class is needed for further improvement. Also, using the metadata of each image can be useful to increase the accuracy of the model.

## 6. References

[1] A. C. Society, "Cancer Facts & Figures 2018," *Atlanta, American Cancer Society,* 2018.
[2] H. K. J. A. o. d. Koh, "Melanoma screening: focusing the public health journey," vol. 143, no. 1, pp. 101-103, 2007.
[3] V. Nikolaou and A. J. B. j. o. d. Stratigos, "Emerging trends in the epidemiology of melanoma," vol. 170, no. 1, pp. 11-19, 2014.
[4] A. C. Society, "Cancer Facts & Figures 2008," *Atlanta, American Cancer Society,* 2008.

[5] H. Safigholi, A. S. Meigooni, and W. Y. J. M. p. Song, "Comparison of 192Ir, 169Yb, and 60Co high-dose rate brachytherapy sources for skin cancer treatment," vol. 44, no. 9, pp. 4426-4436, 2017.
[6] H. Safigholi, W. Y. Song, and A. S. J. J. o. a. c. m. p. Meigooni, "Optimum radiation source for radiation therapy of skin cancer," vol. 16, no. 5, pp. 219-227, 2015.
[7] Z. Ouhib et al., "Aspects of dosimetry and clinical practice of skin brachytherapy: The American Brachytherapy Society working group report," vol. 14, no. 6, pp. 840-858, 2015.
[8] U.-O. Dorj, K.-K. Lee, J.-Y. Choi, M. J. M. T. Lee, and Applications, "The skin cancer classification using deep convolutional neural network," pp. 1-16, 2018.
[9] A. Esteva et al., "Dermatologist-level classification of skin cancer with deep neural networks," vol. 542, no. 7639, p. 115, 2017.
[10] D. Ruiz, V. Berenguer, A. Soriano, and B. J. E. S. w. A. SáNchez, "A decision support system for the diagnosis of melanoma: A comparative approach," vol. 38, no. 12, pp. 15217-15223, 2011.
[11] S. V. Mohan and A. L. S. J. C. d. r. Chang, "Advanced basal cell carcinoma: epidemiology and therapeutic innovations," vol. 3, no. 1, pp. 40-45, 2014.
[12] B. Lindelöf and M. A. J. T. J. o. d. Hedblad, "Accuracy in the clinical diagnosis and pattern of malignant melanoma at a dermatological clinic," vol. 21, no. 7, pp. 461-464, 1994.
[13] C. Morton and R. J. T. B. j. o. d. Mackie, "Clinical accuracy of the diagnosis of cutaneous malignant melanoma," vol. 138, no. 2, pp. 283-287, 1998.
[14] G. Argenziano and H. P. J. T. l. o. Soyer, "Dermoscopy of pigmented skin lesions–a valuable tool for early," vol. 2, no. 7, pp. 443-449, 2001.
[15] M.-L. Bafounta, A. Beauchet, P. Aegerter, and P. J. A. o. d. Saiag, "Is dermoscopy (epiluminescence microscopy) useful for the diagnosis of melanoma?: Results of a meta-analysis using techniques adapted to the evaluation of diagnostic tests," vol. 137, no. 10, pp. 1343-1350, 2001.
[16] M. Vestergaard, P. Macaskill, P. Holt, and S. J. B. J. o. D. Menzies, "Dermoscopy compared with naked eye examination for the diagnosis of primary melanoma: a meta-analysis of studies performed in a clinical setting," vol. 159, no. 3, pp. 669-676, 2008.
[17] G. Salerni et al., "Meta-analysis of digital dermoscopy follow-up of melanocytic skin lesions: a study on behalf of the International Dermoscopy Society," vol. 27, no. 7, pp. 805-814, 2013.
[18] M. Binder et al., "Epiluminescence microscopy: a useful tool for the diagnosis of pigmented skin lesions for formally trained dermatologists," vol. 131, no. 3, pp. 286-291, 1995.
[19] R. P. Braun, H. S. Rabinovitz, M. Oliviero, A. W. Kopf, and J.-H. J. J. o. t. A. A. o. D. Saurat, "Dermoscopy of pigmented skin lesions," vol. 52, no. 1, pp. 109-121, 2005.
[20] H. Kittler, H. Pehamberger, K. Wolff, and M. J. T. l. o. Binder, "Diagnostic accuracy of dermoscopy," vol. 3, no. 3, pp. 159-165, 2002.
[21] D. Piccolo, A. Ferrari, K. Peris, R. Daidone, B. Ruggeri, and S. J. B. J. o. D. Chimenti, "Dermoscopic diagnosis by a trained clinician vs. a clinician with minimal dermoscopy training vs. computer-aided diagnosis of 341 pigmented skin lesions: a comparative study," vol. 147, no. 3, pp. 481-486, 2002.
[22] H. Pehamberger, A. Steiner, and K. J. J. o. t. A. A. o. D. Wolff, "In vivo epiluminescence microscopy of pigmented skin lesions. I. Pattern analysis of pigmented skin lesions," vol. 17, no. 4, pp. 571-583, 1987.

[23]  A. Steiner, H. Pehamberger, and K. J. A. r. Wolff, "Improvement of the diagnostic accuracy in pigmented skin lesions by epiluminescent light microscopy," vol. 7, no. 3 Pt B, pp. 433-434, 1987.

[24]  C. Dolianitis, J. Kelly, R. Wolfe, and P. J. A. o. d. Simpson, "Comparative performance of 4 dermoscopic algorithms by nonexperts for the diagnosis of melanocytic lesions," vol. 141, no. 8, pp. 1008-1014, 2005.

[25]  P. Carli *et al.*, "Pattern analysis, not simplified algorithms, is the most reliable method for teaching dermoscopy for melanoma diagnosis to residents in dermatology," vol. 148, no. 5, pp. 981-984, 2003.

[26]  M. Burroni *et al.*, "Melanoma computer-aided diagnosis: reliability and feasibility study," vol. 10, no. 6, pp. 1881-1886, 2004.

[27]  D. Gutman *et al.*, "Skin lesion analysis toward melanoma detection: A challenge at the international symposium on biomedical imaging (ISBI) 2016, hosted by the international skin imaging collaboration (ISIC)," 2016.

[28]  B. Rosado *et al.*, "Accuracy of computer diagnosis of melanoma: a quantitative meta-analysis," vol. 139, no. 3, pp. 361-367, 2003.

[29]  A. Masood and A. J. I. j. o. b. i. Ali Al-Jumaily, "Computer aided diagnostic support system for skin cancer: a review of techniques and algorithms," vol. 2013, 2013.

[30]  C. Barata, M. E. Celebi, J. S. J. I. j. o. b. Marques, and h. informatics, "Improving dermoscopy image classification using color constancy," vol. 19, no. 3, pp. 1146-1152, 2015.

[31]  R. Garnavi, M. Aldeen, and J. J. I. t. o. i. t. i. b. Bailey, "Computer-aided diagnosis of melanoma using border-and wavelet-based texture analysis," vol. 16, no. 6, pp. 1239-1252, 2012.

[32]  J. Glaister, A. Wong, and D. A. J. I. t. o. b. e. Clausi, "Segmentation of skin lesions from digital images using joint statistical texture distinctiveness," vol. 61, no. 4, pp. 1220-1230, 2014.

[33]  S. Kaya *et al.*, "Abrupt skin lesion border cutoff measurement for malignancy detection in dermoscopy images," in *BMC bioinformatics*, 2016, vol. 17, no. 13, p. 367: BioMed Central.

[34]  E. Almansour and M. A. J. I. I. J. C. S. N. S. Jaffar, "Classification of Dermoscopic skin cancer images using color and hybrid texture features," vol. 16, no. 4, pp. 135-9, 2016.

[35]  Q. Abbas, M. E. Celebi, C. Serrano, I. F. GarcíA, and G. J. P. R. Ma, "Pattern classification of dermoscopy images: A perceptually uniform model," vol. 46, no. 1, pp. 86-97, 2013.

[36]  G. Capdehourat, A. Corez, A. Bazzano, R. Alonso, and P. J. P. R. L. Musé, "Toward a combined tool to assist dermatologists in melanoma detection from dermoscopic images of pigmented skin lesions," vol. 32, no. 16, pp. 2187-2196, 2011.

[37]  I. Giotis, N. Molders, S. Land, M. Biehl, M. F. Jonkman, and N. J. E. s. w. a. Petkov, "MED-NODE: a computer-assisted melanoma diagnosis system using non-dermoscopic images," vol. 42, no. 19, pp. 6578-6585, 2015.

[38]  A. G. Isasi, B. G. Zapirain, A. M. J. C. i. B. Zorrilla, and Medicine, "Melanomas non-invasive diagnosis application based on the ABCD rule and pattern recognition image processing algorithms," vol. 41, no. 9, pp. 742-755, 2011.

[39]  Y. LeCun, Y. Bengio, and G. J. n. Hinton, "Deep learning," vol. 521, no. 7553, p. 436, 2015.

[40] A. Krizhevsky, I. Sutskever, and G. E. Hinton, "Imagenet classification with deep convolutional neural networks," in *Advances in neural information processing systems*, 2012, pp. 1097-1105.

[41] L. Yu, H. Chen, Q. Dou, J. Qin, and P.-A. J. I. t. o. m. i. Heng, "Automated melanoma recognition in dermoscopy images via very deep residual networks," vol. 36, no. 4, pp. 994-1004, 2017.

[42] H. Haenssle *et al.*, "Man against machine: diagnostic performance of a deep learning convolutional neural network for dermoscopic melanoma recognition in comparison to 58 dermatologists," vol. 29, no. 8, pp. 1836-1842, 2018.

[43] S. S. Han, M. S. Kim, W. Lim, G. H. Park, I. Park, and S. E. J. J. o. I. D. Chang, "Classification of the Clinical Images for Benign and Malignant Cutaneous Tumors Using a Deep Learning Algorithm," 2018.

[44] P. Tschandl, C. Rosendahl, and H. J. a. p. a. Kittler, "The HAM10000 Dataset: A Large Collection of Multi-Source Dermatoscopic Images of Common Pigmented Skin Lesions," 2018.

[45] T. Mendonça, P. M. Ferreira, J. S. Marques, A. R. Marcal, and J. Rozeira, "PH 2-A dermoscopic image database for research and benchmarking," in *Engineering in Medicine and Biology Society (EMBC), 2013 35th Annual International Conference of the IEEE*, 2013, pp. 5437-5440: IEEE.

[46] O. Russakovsky *et al.*, "Imagenet large scale visual recognition challenge," vol. 115, no. 3, pp. 211-252, 2015.

[47] M. Abadi *et al.*, "Tensorflow: a system for large-scale machine learning," in *OSDI*, 2016, vol. 16, pp. 265-283.

[48] C. Szegedy, V. Vanhoucke, S. Ioffe, J. Shlens, and Z. Wojna, "Rethinking the inception architecture for computer vision," in *Proceedings of the IEEE conference on computer vision and pattern recognition*, 2016, pp. 2818-2826.

[49] C. Szegedy, S. Ioffe, V. Vanhoucke, and A. A. Alemi, "Inception-v4, inception-resnet and the impact of residual connections on learning," in *AAAI*, 2017, vol. 4, p. 12.

[50] K. He, X. Zhang, S. Ren, and J. Sun, "Deep residual learning for image recognition," in *Proceedings of the IEEE conference on computer vision and pattern recognition*, 2016, pp. 770-778.

[51] G. Huang, Z. Liu, L. Van Der Maaten, and K. Q. Weinberger, "Densely Connected Convolutional Networks," in *CVPR*, 2017, vol. 1, no. 2, p. 3.

[52] M. A. Al-antari, M. A. Al-masni, M.-T. Choi, S.-M. Han, and T.-S. J. I. J. o. M. I. Kim, "A fully integrated computer-aided diagnosis system for digital X-ray mammograms via deep learning detection, segmentation, and classification," vol. 117, pp. 44-54, 2018.

[53] G. Litjens *et al.*, "A survey on deep learning in medical image analysis," vol. 42, pp. 60-88, 2017.

[54] A. Garcia-Garcia, S. Orts-Escolano, S. Oprea, V. Villena-Martinez, and J. J. a. p. a. Garcia-Rodriguez, "A review on deep learning techniques applied to semantic segmentation," 2017.